\documentclass{article}
\usepackage{spconf,amsmath,graphicx}

\usepackage{amsmath,amssymb} 
\usepackage{color}
\usepackage{balance}
\usepackage{amsmath}
\usepackage[linesnumbered,ruled,vlined]{algorithm2e}
\usepackage{algpseudocode}
\usepackage{enumitem}


\usepackage{graphicx}
\usepackage{pdfpages}
\usepackage{ragged2e}

\graphicspath{{figures/}} 
\newcommand{\etal}{\textit{et al}.}

\newcommand{\eg}{\textit{e}.\textit{g}.}

\title{Domain Adaptation on Semantic Segmentation for Aerial Images}
\name{Ying Chen, Xu Ouyang, Kaiyue Zhu, Gady Agam}
\address{Illinois Institute of Technology\\
				Department of Computer Science \\
				Chicago, IL 60616, USA\\
				ychen245, xouyang3, kzhu6@hawk.iit.edu, agam@iit.edu}
\begin{document}
%
\maketitle
\begin{abstract}
Semantic segmentation has achieved significant advances in recent years.
While deep neural networks perform semantic segmentation well, their success rely on pixel level supervision which is expensive and time-consuming.
Further, training using data from one domain may not generalize well to data from a new domain due to a domain gap between data distributions in the different domains. This domain gap is particularly evident in aerial images where visual appearance depends on the type of environment imaged, season, weather, and time of day when the environment is imaged. Subsequently, this distribution gap leads to severe accuracy loss when using a pretrained segmentation model to analyze new data with different characteristics. 
In this paper, we propose a novel unsupervised domain adaptation framework to address domain shift in the context of aerial semantic image segmentation. To this end, we solve the problem of domain shift by learn the soft label distribution difference between the source and target domains. Further, we also apply entropy minimization on the target domain to produce high-confident prediction rather than using high-confident prediction by pseudo-labeling. We demonstrate the effectiveness of our domain adaptation framework using the challenge image segmentation dataset of ISPRS, and show improvement over state-of-the-art methods in terms of various metrics.
\end{abstract}

\section{Introduction}

Semantic segmentation is a core image analysis operation where each pixel is assigned a semantic label attributing it to a certain object. In practice, semantic image segmentation normally serves as high-level pre-processing step to support complete scene understanding. In the context of remote sensing, semantic image segmentation is viewed as pixel-wise classification and has been widely studied in a variety of potential applications concerning aerial images, such as monitoring and planning urban areas, disaster detection, and fast emergency response. The growing use of Unmanned Aerial Vehicles (UAVs) increase the need for semantic segmentation of high resolution images to support automatic monitoring of inhabited areas.

\begin{figure*}[h]

\begin{minipage}[b]{1.0\linewidth}
  \centering
  \centerline{\includegraphics[width=\textwidth,height=6cm]{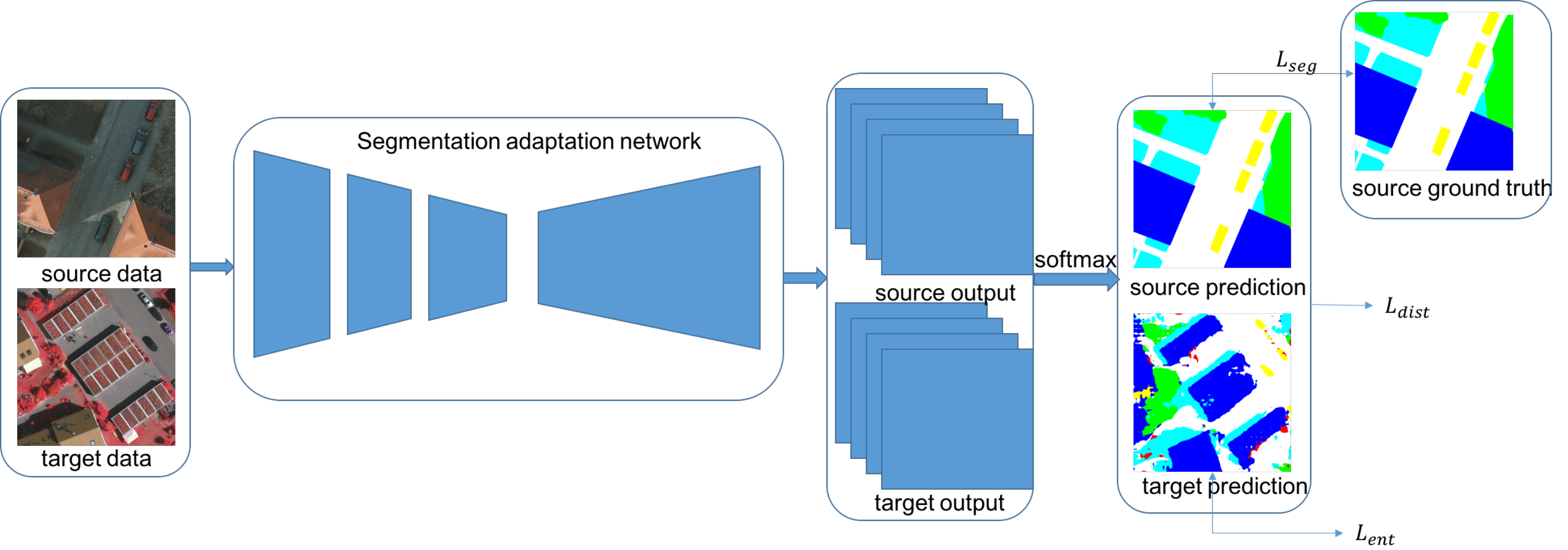}}
\end{minipage}
\caption{Proposed network architecture and loss. $\mathcal{L}_{ent}$ minimizes the entropy of the target sample $x_{t}$. While $\mathcal{L}_{dist}$ is used to enforce the class-wise distribution consistency between source domain and target domain.}
\label{fig:arc}
\end{figure*}

CNN-based fully-supervised approaches have achieved remarkable progress in semantic segmentation on datasets such as Cityscapes~\cite{cordts2016cityscapes} and PASCAL VOC-2012~\cite{everingham2015pascal}. In practice, given sufficient labeled data, training a state-of-the-art network can easily achieve an accuracy over $80\%$~\cite{lateef2019survey}. This high accuracy, however, requires a large labeled dataset for the target domain. 

Producing dense pixel-level annotations for real-world datasets is extremely costly and requires an enormous amount of manual work. For example, the annotation of a single Cityscapes image takes nearly $90$ minutes on average~\cite{richter2016playing}. Using a small training set results in poor generalization from training/source data to test/target data when samples of the target domain have different characteristics. One straightforward way to address this challenge is to apply weakly and semi-supervised methods~\cite{souly2017semi,wei2018revisiting}, where only a small subset of the data is annotated, or where only weak supervision is provided. The application of such approaches may still require time-consuming labeling and weak annotation may be hard to obtain for most real-world applications. Thus, there is a need for efficient ways to address semi-supervised or unsupervised domain adaptation that require a small subset of target domain labeled data or even no target domain labeled data. In this work, we focus on unsupervised domain adaptation.

Unsupervised domain adaptation (UDA) aims at learning a target domain model from a well performing source model. UDA tends to reduce or remove cross-domain shift between the source and target domains by learning a mapping function between them. Most recent UDA methods address this cross-domain shift through minimizing the distribution difference between intermediate features or final outputs of the two domains. Some approaches~\cite{yan2017mind,long2015learning,long2016unsupervised} use maximum mean discrepancies (MMD) or adversarial training~\cite{hoffman2016fcns,tzeng2017adversarial,tsai2018learning,vu2019advent} to solve UDA, while other methods apply self-training~\cite{zou2018unsupervised,li2019bidirectional} to produce pseudo labels, or apply generative networks~\cite{hoffman2018cycada,wu2018dcan,benjdira2019unsupervised,benjdira2020data} to generate target data. Semi-supervised learning solves a problem related to domain adaptation and many of its strategies can be used to address domain adaptation, such as self-training, and entropy minimization~\cite{vu2019advent,li2019bidirectional}. 


In this work, we propose a new framework for domain adaptation of semantic segmentation in aerial images. An overview of the proposed network is shown in Figure\ref{fig:arc}, and is detailed in Section \ref{sec:proposed-approach}. Our approach involves two strategies: First, we employ entropy minimization to UDA following an observation that models trained on labeled source domain and unlabeled target domain tend to get over-confident/low-entropy and under-confident/high-entropy results, respectively~\cite{vu2019advent}. Thus, to address this we use entropy minimization on the target domain to reduce its entropy. Second, following an assumption that the data distribution for each class should be the same when applying domain adaptation, we use the Kullback-Leibler divergence (KL-divergence)~\cite{kullback1951information} to address the class-wise distribution difference between the source and target domains, instead of attempting to align pixel-level feature distributions. To the best of our knowledge, we are the first to apply class-wise distribution alignment to address domain adaptation in semantic segmentation.

The key contributions of this paper are as follows: 
\begin{itemize}
 \item We propose a simple yet effective end-to-end approach. In contrast to common methods, which are 
adversarial and are known to be difficult to train~\cite{liu2017approximation}, the proposed method is not adversarial.
  \item We employ entropy minimization to improve low-confidence predictions on the target domain for semantic segmentation of aerial images. 
  \item We propose aligning class-wise distributions of source and target domains using KL-divergence loss to support appearance differences between the source and target domains.
   \item We demonstrate that our approach is able to mitigate domain shift between source and target domains and surpass state-of-the-art methods when evaluated on a standard ISPRS segmentation challenge dataset.
\end{itemize}

\section{Related work}

This section summarizes related work for domain adaptation in semantic segmentation. 
The domain adaptation approach we propose in this paper is non-generative and combines ideas from both unsupervised and self-supervised approaches. 
Subsequently, we review unsupervised domain adaptation (both non-generative and generative) and self-supervised domain adaption approaches.
Relevant approaches in this review are used for comparison in the experimental evaluation section.
To the best of our knowledge, we are the first to apply entropy based UDA for semantic segmentation. 

\subsection{Unsupervised domain adaptation.}
Domain adaptation aims at helping a trained model better generalize to unseen test data. Numerous unsupervised domain adaptation (UDA) methods~\cite{long2015learning,long2017deep,long2016unsupervised,sohn2017unsupervised,xie2018learning} for image classification and detection have been proposed to address domain shift between labeled source and unlabeled target domains. The main idea behind current approaches is to minimize distribution discrepancy between source and target data. The main approaches for tackling this distribution discrepancy can be classified as global distribution alignment, and class-wise or conditional distribution alignment. Recent work include maximum mean discrepancies (MMD)~\cite{long2015learning,long2016unsupervised}, adversarial training~\cite{tzeng2017adversarial,tsai2018learning,vu2019advent}, or self-training with pseudo labels~\cite{zou2018unsupervised,li2019bidirectional}. Most of existing UDA methods are adversarial.

\subsection{Non-Generative UDA for semantic segmentation.}
Recent work concerning UDA for semantic segmentation use synthetic data as the source domain (\eg~ SYNTHIA~\cite{ros2016synthia}). Adversarial training is the most common method for UDA for semantic segmentation. It uses two networks. One network aims to predict the segmentation map for labeled training data from either the source domain or target domain, while the other network is used to align feature distribution between the two domains taken from the segmentation network in an adversarial manner. 
The first unsupervised domain adaptation method for transferring semantic segmentation across image domains was proposed by Hoffman \etal~\cite{hoffman2016fcns}, where they use global and category specific adaptation techniques to combine global and local alignment methods. Chen \etal~\cite{chen2017no} propose a similar approach to perform global and class-wise adaptation, and use adversarial learning for assigning pseudo labels to achieve joint global and class-wise adaptation of segments. 

In~\cite{chen2018road}, adversarial training is used to adapt from synthetic to real urban scenes by using spatial-aware adaptation loss along with a distillation loss. In this approach, feature maps are divided into multiple grids and a MMD loss is calculated in each grid cell. Hong \etal~\cite{hong2018conditional} propose a principled approach to model the residual in the feature space between the source and target domains while maintaining their semantic spatial layouts. Similarly, Tsai \etal~\cite{tsai2018learning} propose an adversarial learning approach to benefit from spatial similarities between the source and target domains, while performing domain adaptation at different feature levels. Vu \etal~\cite{vu2019advent} propose an adversarial entropy minimization approach to address domain shift between source and target domains with loss based on the entropy of pixel-wise predictions. Their work involves two complementary methods using an entropy loss and an adversarial loss. Considering the fact that the gradient of the entropy is biased towards samples that are easy to transfer, a similar approach~\cite{chen2019domain} replaces the entropy loss with a maximum squares loss. 

\subsection{Generative UDA for semantic segmentation.}
Generative networks can be used for domain adaptation in semantic segmentation by conditioning target images on the source domain. Hoffman \etal~\cite{hoffman2018cycada} use a Cycle-Consistent Adversarial Domain Adaptation (CyDADA) model to perform both pixel-level adaptation and feature-level representation adaptation. They use the CycleGAN~\cite{zhu2017unpaired} network to produce a target image conditioned on a source image. To preserve spatial structures and semantic information, Wu \etal~\cite{wu2018dcan} perform channel-wise feature alignment in both an image generator and a segmentation network. The generator produces samples from the target domain with content from the source domain while maintaining the style of the target domain. Li \etal~\cite{li2019bidirectional} propose a bidirectional learning framework to alternate learning image translation using CycleGAN. Similarly, the authors  in~\cite{benjdira2019unsupervised} and~\cite{benjdira2020data} employ CycleGAN based image translation to translate the target domain to the source domain and increase the ability of the model to work on the new target domain. 

\subsection{Self-supervised DA for semantic segmentation.}
Another category of approaches addressing UDA involves self-training. The core idea is to generate pseudo labels for unlabeled target domain data using an ensemble of previous models. Zou \etal~\cite{zou2018unsupervised} propose an iterative self-training (ST) procedure to generate pseudo labels for target data and use them to retrain the model. Class balancing is used to avoid gradual dominance, and spatial priors are used to refine the generated labels. In~\cite{li2019bidirectional}, self-training is employed to learn a better segmentation model and in return improve an image translation model. 

Entropy minimization was shown to be useful for addressing semi-supervised learning problems~\cite{grandvalet2005semi,springenberg2015unsupervised}, and was used to solve domain adaptation for classification~\cite{long2016unsupervised} and semantic segmentation of natural images~\cite{vu2019advent}.

\section{Proposed approach}
\label{sec:proposed-approach}
In this section, we present the proposed framework of domain adaptation for semantic segmentation in aerial images, where we use a combination of entropy minimization and soft class-wise distribution alignment. 
\subsection{Supervised semantic segmentation}
Our domain adaptation approach relies on a single segmentation adaptation network. An overview of the proposed model is provided in Figure~\ref{fig:arc}.

The model is first learned on the source domain using a supervised segmentation loss. We denote the labeled source domain as $D_{s}=\{(x_{s},y_{s})|x_{s}\in {R^{H \times W \times 3}}, y_{s} \in {(1,C)^{H \times W}}\}$, where each sample is an $H \times W$ color image which is associated with a ground-truth C-class segmentation map. Each entry $y_s^{i,j}$ takes a class label from a finite set (${1,2,...,C}$) or a one-hot vector $[y_{s}^{(i,j,c)}]_{c}$. Similarly, we denote the unlabeled target domain using $D_{t}=\{(x_{t})|x_{t}\in {R^{H \times W \times 3}}\}$. We forward the source image $x_{s}$  to a semantic segmentation network $F$. The network output is a segmentation map $F(x_{s})$ with dimension $H \times W \times C$. After passing through a softmax layer, we predict the segmentation softmax output $P_{x_{s}}=[P_{x_{s}}^{(i,j,c)}]_{i,j,c}=\mbox{softmax}(F(x_{s}))$, where each C-dimensional vector is normalized by the softmax to represent a discrete distribution over classes. The model $F$ is trained on source data in a supervised manner using the categorical cross entropy loss:
\begin{equation}
\mathcal{L}_{ce}(x_{s},y_{s})=-\dfrac{1}{N}\sum_{i=1}^{N} \sum_{j=1}^{C} y_{s}^{(i,j)}\log P_{x_{s}}^{(i,j)}
\label{eq: seg}
\end{equation}
where $N = H \times W$ is the total number of pixels in an image.

\subsection{Entropy minimization}
In our UDA approach, the target domain is unlabeled, and so supervised learning based on the source domain may result in producing over-confident predictions for source examples, and under-confident predictions for target examples. To improve the test performance on target data, we could use a self-supervised-learning (SSL) method to select target pixel predictions with sufficient confidence and use them as pseudo labels for target samples during training. Specifically, based on the prediction probability produced for the target domain, we could use SSL to obtain pseudo labels $\hat{y}_{t}$ with high confidence, using a fixed or scheduled threshold. The categorical cross entropy loss on target predictions using these pseudo labels is given by:
\begin{equation}
\mathcal{L}_{ce}(x_{t},\hat{y}_{t})=-\dfrac{1}{N}\sum_{i=1}^{N} \sum_{j=1}^{C} \hat{y}_{t}^{(i,j)}\log P_{x_{t}}^{(i,j)}
\label{eq: seg1}
\end{equation}

In contrast to SSL where a threshold needs to be selected for hard assignment, in the proposed approach we use soft assignment. Specifically, we use the entropy minimization method of~\cite{vu2019advent} to constrain the model to directly produce high confidence predictions, by minimizing the Shannon entropy~\cite{shannon1948mathematical} of target domain predictions. 

To produce target domain predictions with high-confidence/certainty, we minimize the entropy loss $\mathcal{L}_{ent}$ of~\cite{vu2019advent} which is given by:

\begin{equation}
\mathcal{L}_{ent}(x_{t})=-\dfrac{1}{N}\sum_{i=1}^{N} \dfrac{1}{\log(C)}\sum_{j=1}^{C}P_{x_{t}}^{(i,j)}\log P_{x_{t}}^{(i,j)}
\label{eq: ent}
\end{equation}

By comparing $\mathcal{L}_{ent}(x_t)$ and $\mathcal{L}_{ce}(x_{t},\hat{y}_{t})$, we observe that these two loss terms are nearly equivalent except that $\mathcal{L}_{ent}(x_t)$ is a soft-assignment version of $\mathcal{L}_{ce}(x_{t},\hat{y}_{t})$. Using a soft assignment in the proposed approach avoids the need for threshold selection in SSL which may be unstable.

\subsection{Learning soft label distribution}
By applying entropy minimization, the proposed approach is able to produce low-entropy when trained on unlabeled target data. However, there may still be a distribution difference between the source and target domains. To address this, we employ a method for class-wise distribution alignment. The approach is illustrated in algorithm~\ref{algo: da}. We first apply the softmax function to the logits $F(x_{s})$ and $F(x_{t})$ produced by the network $F$ for source data and target data respectively. We then add all the values of $P_{x_{s}}$ and $P_{x_{t}}$ via the channel dimension to produce C-dimensional vectors, and calculate soft counts for each class in both domains. Finally, by performing normalization on the vectors, we obtain soft class-wise distributions for source and target domains. 

\begin{algorithm}
\begin{enumerate}[leftmargin=0pt,rightmargin=\parindent,itemindent=0.5em]
\justifying
\item Generate the number of classes C-dimensional logits/soft-segmentation maps $F(x_{s})$ and $F(x_{t})$.
\item Produce probability maps $P_{x_{s}}$ and $P_{x_{t}}$ using softmax.
\item In each channel, add all the values to produce C-dimensional soft count vectors $[P_{x_{s}}^{c}]_{c}$ and $[P_{x_{t}}^{c}]_{c}$.
\item Normalize $[P_{x_{s}}^{c}]_{c}$ and $[P_{x_{t}}^{c}]_{c}$ to produce class-wise data distributions $\hat{P}_{x_{s}}$ and $\hat{P}_{x_{t}}$ for both domains.
\item Minimize the loss $\mathcal{L}_{dist}(x_{s},x_{t})$ in equation~\ref{eq: seg2} through backpropagation to align class-wise distribution.
\end{enumerate}
\caption{Class-wise distribution alignment between source and target data}
\label{algo: da}
\end{algorithm}

After obtaining the soft class-wise distributions $\hat{P}_{x_{s}}$ and $\hat{P}_{x_{t}}$ for the source and target domains respectively, we compute the KL-divergence loss to measure the distance between the two distributions:
%
\begin{equation}
\mathcal{L}_{dist}(x_{s},x_{t})=-\dfrac{1}{C}\sum_{c=1}^{C} \hat{P}_{x_{s}}\log \dfrac{\hat{P}_{x_{s}}}{\hat{P}_{x_{t}}}
\label{eq: seg2}
\end{equation}
We add this loss term to the combined network loss and so force the source and target distributions to be similar.

\subsection{Objective function for UDA in aerial images}
During training, the model is learned by jointly minimizing the supervised segmentation loss $\mathcal{L}_{seg}$ on source data, the entropy loss $\mathcal{L}_{ent}$ on target data, and the distribution similarity loss $\mathcal{L}_{dist}$. The overall loss function for UDA of semantic segmentation is given by:

\begin{multline}
\mathcal{L}=\min\limits_{\theta_{F}}[\dfrac{1}{\left|D_{s}\right|}\sum\limits_{x_{s}}\mathcal{L}_{seg}(x_{s},y_{s}) + \dfrac{\lambda_{1}}{\left|D_{t}\right|}\sum\limits_{x_{t}}\mathcal{L}_{ent}(x_{t}) + \\
\dfrac{\lambda_{2}}{\left|D\right|} \sum\limits_{(x_{s},x_{t})}\mathcal{L}_{dist}(x_{s},x_{t})]
\label{eq: loss}
\end{multline}
where $D$ denotes $min(|D_{s}|,|D_{t}|)$, and $\lambda_{1}$, $\lambda_{2}$ control the tradeoff among the three loss terms.

\subsection{Network architecture and training}
In this section, we combine the components described above. The network architecture is illustrated in Figure~\ref{fig:arc}.
\subsubsection{Segmentation network architecture.}
To obtain high-quality segmentation results, it is critical to choose a strong baseline model. In this work, we use Deeplab-V2~\cite{chen2017deeplab} with a pretrained ResNet-101~\cite{he2016deep} model as the base semantic segmentation network $F$. Similar to several recent works on UDA~\cite{tsai2018learning,vu2019advent}, we remove the multi-scale fusion strategy due to limited memory, and remove the last classification layer which is of no need for our problem. To better capture the scene context, we apply Atrous Spatial Pyramid Pooling (ASPP)~\cite{chen2017deeplab} as the final classifier after the last layer. Following the setting in~\cite{chen2017deeplab}, the sampling rates are set to $\{6, 12, 18, 24\}$. We also change the strides of the last layers $conv4$ and $conv5$ using dilated convolution layers to enlarge the receptive field. Finally, an up-sampling layer is applied to match the size of the input. The segmentation outputs of source and target inputs are passed to the three loss terms.

\subsubsection{Network training.}

To train the model, we first process the source input to optimize the segmentation network using $\mathcal{L}_{seg}$ and output the segmentation prediction $P_{x_{s}}$. Likewise, we generate the output $P_{x_{t}}$ for the target input, and pass it to optimize $\mathcal{L}_{ent}$. Finally, after calculating the soft counts per class for $P_{x_{s}}$ along with $P_{x_{t}}$, we obtain $\hat{P}_{x_{s}}$ and $\hat{P}_{x_{t}}$ which are passed to optimize $\mathcal{L}_{dist}$.

We use the Pytorch deep learning framework~\cite{paszke2017automatic} to implement our network on two NVIDIA-SMI GPU with $16$ GB memory in total. We use a Stochastic Gradient Descent~\cite{bottou2010large} optimizer with momentum of $0.9$ and a weight decay of $10^{-4}$ to train the model. The initial learning rate is set to $0.001$ and decayed using the polynomial decay schedule of~\cite{he2016deep}. The input size for the source and target domains is given by $512\times512$, and the batch size is set as $2$. The two loss weight factors $\lambda_{1}$ and $\lambda_{2}$  are set to $0.001$ and $0.1$, respectively. 

\begin{figure}[h]

\begin{minipage}[b]{1.0\linewidth}
  \centering
  \centerline{\includegraphics[width=\textwidth,height=8cm]{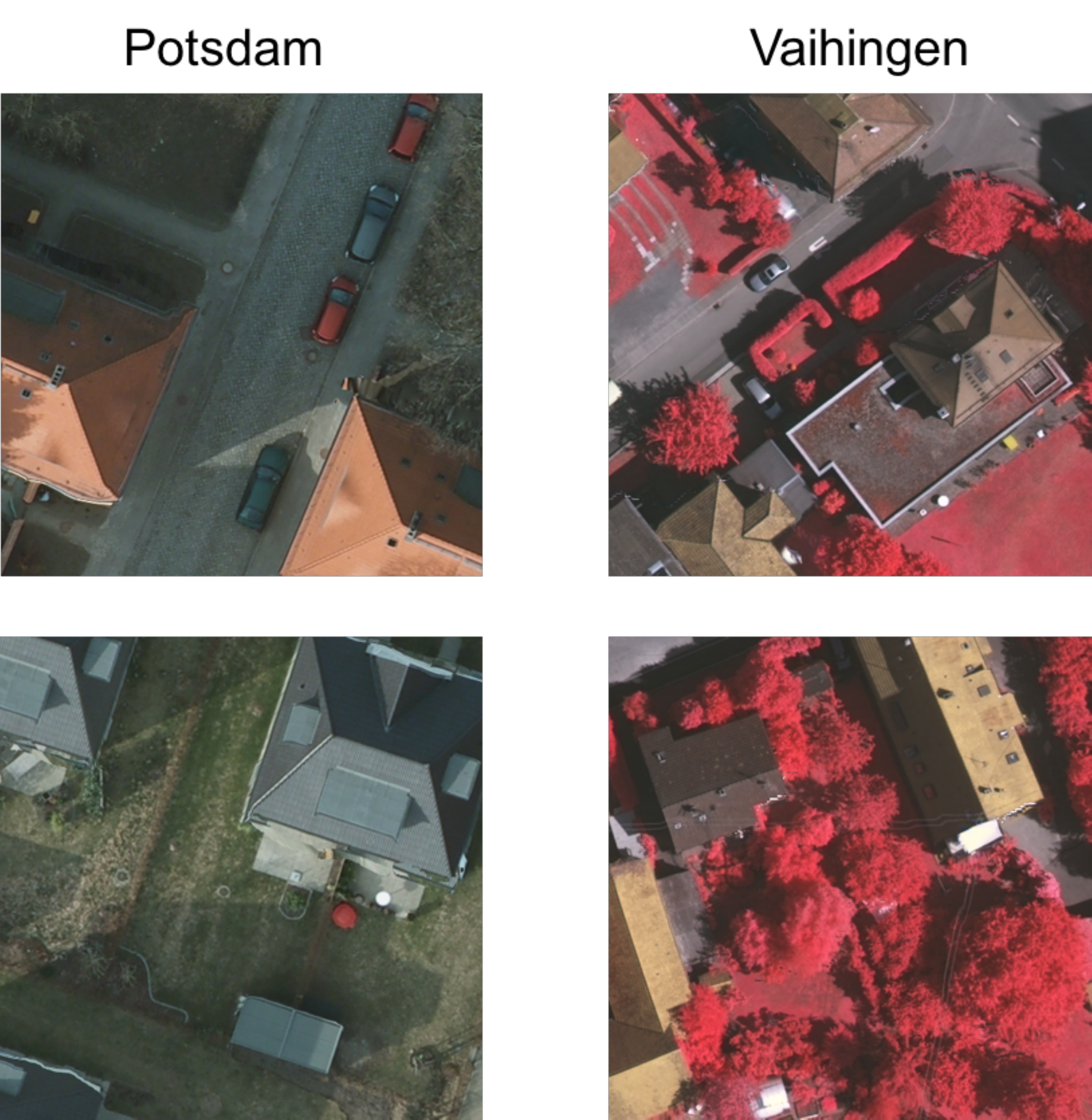}}
\end{minipage}
\caption{Example images from the Potsdam and Vaihingen datasets.}
\label{fig:sample}
\end{figure}

\section{Experiments}
In this section, we present experimental results using several evaluation metrics. We evaluate the proposed approach and compare it to known approaches using standard evaluation datasets.

 \begin{table*}
\caption{Evaluation results using the Vaihingen target domain. The symbol $*$ indicates data not provided in reference paper. DBDA is the proposed approach and DBDA$\dag$ is a variant of the proposed approach provided for ablation study.}
\begin{center}
\begin{tabular}{|l|l||ccccc|} 
  \hline
  Category & Method & Average Accuracy & Precision  & Recall & F1-score & IoU\\
  \hline
  &Source-only         &0.471  &0.414 &0.471 &0.316 &0.214\\
  &FCNs in wild~\cite{hoffman2016fcns}      &0.486  &* &* &* &0.309\\
  Baseline&AUDA~\cite{benjdira2019unsupervised} &0.520  &0.540 &0.520 &0.490 &0.300 \\
  approaches&DUDA~\cite{benjdira2020data}      &0.588  &* &* &* &0.349\\
  &AdvEnt~\cite{vu2019advent}      &0.308  &0.333 &0.308 &0.202 &0.129\\
  &MinEnt~\cite{vu2019advent}      &0.329  &0.383 &0.329 &0.236 &0.237\\
  \hline
  Proposed& DBDA$\dag$  &0.554  &0.546 &0.554 &0.510 &0.367\\
  approach&  DBDA &\textbf{0.591}  &\textbf{0.568} &\textbf{0.591} &\textbf{0.539}&\textbf{0.393}\\
  \hline
\end{tabular}
\label{table:metrics}
\end{center}
\end{table*}

\subsection{Datasets}
To evaluate the proposed method, we follow a common evaluation procedure as described in~\cite{benjdira2019unsupervised,benjdira2020data}. To strengthen this evaluation procedure we add an additional task of adapting from the target set into the source set (in addition to adapting from the source set to the target set). 
We use the ISPRS (WGII/4) 2D semantic segmentation benchmark dataset~\cite{labeling2016use}, which is part of the ISPRS 2D semantic labeling challenge. We are specifically interested in aerial image segmentation with its unique challenges stemming from appearance differences due to sensors, seasonal changes, illumination changes, and other natural phenomena, and which are represented well in this dataset. 

Following the standard evaluation procedure 
we use two cities, Potsdam and Vaihingen, as our source and target domains respectively. We use images from the original digital surface model (DSM) for our domain adaptation task. The resolution in these  datasets, while high, does not match. Potsdam has a resolution of $5$cm per pixel whereas Vaihingen has a resolution of $7$cm per pixel.  
This resolution difference contributes to one aspect of domain shift between the Potsdam source and Vaihingen target datasets. As stated in~\cite{benjdira2020data}, this set has two more domain shift factors having to do with sensor variation and variation of class representations.
As stated earlier, in addition to the standard evaluation where we use Potsdam and Vaihingen as source 
and target domains respectively, we conduct an additional evaluation where we reverse their roles. This in essence doubles our evaluation set.

Each pixel in each image is annotated with one of six classes of ground truth objects: building, tree, low vegetation, car, clutter/background, and impervious surfaces. The Potsdam dataset contains $38$ TOP images of size $6000 \times 6000$, whereas the Vaihingen dataset contains $33$ TOP images with the same size. Following the split criterion in~\cite{benjdira2019unsupervised,benjdira2020data}, each image is divided into several squares of size $512$ by $512$ for both training and testing. 32 images of the Potsdam dataset are used for training and the remaining six are used for testing. In the Vaihingen dataset $27$ out of $33$ images are used for training and the remaining $6$ for testing. Several examples from the source domain (Potsdam) and the target domain (Vaihingen) are shown in Figure~\ref{fig:sample}. The distribution of pixels over the six classes is presented in~\cite{benjdira2020data}, where it is shown that the class distributions for the two domains are matched well. The class distribution in each dataset is not balanced.

\begin{table*}
\caption{Evaluation results when reversing the roles of source and target domains (using Potsdam as target). Fewer baseline methods are reported compared with Table \ref{table:metrics} because some of the compared methods do not report this reverse evaluation.}
\begin{center}
\begin{tabular}{|l|l||ccccc|} 
  \hline
  Category & Method & Average Accuracy & Precision  & Recall & F1-score & IoU\\
  \hline
  &Source-only         &0.396  &0.512 &0.396 &0.372 &0.243\\
  Baseline &DUDA~\cite{benjdira2020data}      &0.363  &0.365 &0.363 &0.318 &0.309 \\
  &AdvEnt~\cite{vu2019advent}      &0.369   &0.548 &0.369 &0.382 &0.268 \\
  &MinEnt~\cite{vu2019advent}      &0.362  &0.436 &0.362 &0.362 &0.246\\
  \hline
  Proposed& DBDA$\dag$  &0.451  &0.582 &0.451 &0.479 &0.319\\
 approach &DBDA &\textbf{0.455}  &\textbf{0.583} &\textbf{0.455} &\textbf{0.485}&\textbf{0.330}\\
  \hline
\end{tabular}
\label{table:metricsba}
\end{center}
\end{table*}

\subsection{Evaluation metrics}
To evaluate the proposed method, we use five commonly used metrics following baseline methods~\cite{benjdira2020data,benjdira2019unsupervised}: accuracy, recall, precision, F1-score, and Intersection over Union (IoU). The metrics are calculated with true positive (TP), false positive (FP), true negative (TN), and false negative (FN). For semantic segmentation, TP and TN measure the number of pixels correctly predicted, whereas FP and FN measure the number of pixels incorrectly predicted. Since we have six different classes, the metrics are calculated as the mean value over the values calculated for all classes separately.  
%

\begin{table*}
\caption{IoU evaluation results for every class on the Vaihingen target domain.}
\begin{center}
\begin{tabular}{|l|l||ccccccc|} 
   \hline
   Category & Method & Imp.Sur. & Bui.  & Lo. Veg. &Tree & Car &Clu.Backgr. &mIoU\\
  \hline
  Baseline &Source-only         &0.413  &0.074 &0.022 &0.306 &0.019 &0.451 &0.214\\ 
  &AdvEnt~\cite{vu2019advent}      &0.414  &0.024 &0.039 &0.013 &0.232 &0.052 &0.129\\
  &MinEnt~\cite{vu2019advent}      &0.461  &0.096 &0.023 &0.003 &0.080 &0.253 &0.237\\
\hline
  Proposed &DBDA$\dag$   &0.546  &0.008 &0.266 &0.523 &0.244 &0.545 &0.367\\
  & DBDA &\textbf{0.560}  &\textbf{0.106} &\textbf{0.312} &\textbf{0.528}&\textbf{0.266} &\textbf{0.589} &\textbf{0.393}\\
  \hline
\end{tabular}
\label{table:ious}
\end{center}
\end{table*}

\subsection{Results}

In this section, we report experimental results of our proposed distribution-based domain adaptation approach (DBDA) and compare it with different baseline approaches. We show that the proposed DBDA approach improves on state-of-the-art performance when evaluated on the ISPRS aerial benchmark. The baseline methods we compare with include: Source-only, a method trained on only the source domain (without domain adaptation) using the segmentation network of Deeplab-V2 (ResNet-101 which is used a backbone for all baseline methods to make the comparison fair); FCNs in wild~\cite{hoffman2016fcns}, a domain adaptation method for semantic segmentation;  AUDA~\cite{benjdira2019unsupervised}, an unsupervised domain adaptation approach for semantic segmentation of aerial images; DUDA~\cite{benjdira2020data}, another domain adaptation algorithm for semantic segmentation of aerial images; AdvEnt~\cite{vu2019advent}, an adversarial domain adaptation approach which directly minimizes the entropy between the target and source domains; and MinEnt~\cite{vu2019advent} which uses entropy loss to directly maximize the prediction certainty in the target domain. 

In addition, we test a variation of our proposed DBDA network termed DBDA$\dag$ in which we remove the entropy loss of the target domain to verify the effectiveness of the entropy loss in our approach. Among these methods, FCNs in wild, AUDA , AdvEnt, and DUDA are all adversarial learning methods.

Table~\ref{table:metrics} presents the evaluation results of semantic segmentation on the Vaihingen test data in terms of the five measures: mean accuracy, precision, recall, F1-score and IoU. As can be observed, our proposed DBDA$\dag$ approach, obtains better performance compared with the baseline approaches in almost all measures (including the source-only approach which does not perform domain adaptation). It validates the advantage of using class-wise distribution alignment to force the target domain to have similar class-wise data distribution as the source domain, which addresses domain shift between the two domains. By adding the minimum entropy component to the proposed approach we observe that the proposed DBDA approach surpasses all other methods.
This demonstrates that our proposed method addresses the issue of domain shift and guarantees high prediction certainty on target data. In addition, our method does not involve any adversarial training as used for image translation in AUDA~\cite{benjdira2019unsupervised} and DUDA~\cite{benjdira2020data}, and so is easier to train.
%

The poor performance of AdvEnt on the test set demonstrates that using adversarial training to adapt domain shift may not work well on aerial data where there are large difference of appearance between the different domains. 
The poor performance of MinEnt on the test set indicates that domain discrepancy remains high even though the prediction certainty is maximized in the target domain. 


To further demonstrate the effectiveness of our method, we perform an additional evaluation where we reverse the role of the source and target domains. In this additional evaluation we use Vaihingen as source and Potsdam as target. Table~\ref{table:metricsba} shows the evaluation performance. As can be observed in the table, the proposed DBDA approach outperforms the baseline methods in all metrics. This demonstrates again that the proposed approach effectively addresses domain shift between the two domains. 
We also observe that AdvEnt and MinEnt show better performance than DUDA for most metrics, which validates the usage of entropy minimization to achieve high prediction certainty in the target domain.


Given that IoU is the most common measure used to evaluate semantic segmentation, we measure per class IoU performance of the proposed approach and the compared methods. The results are provided in Table~\ref{table:ious}. As shown in the table, the proposed DBDA$\dag$  approach gets improvements on almost every class, 
which demonstrates that DBDA$\dag$ is able to reduce domain shift between the two domains. Our full DBDA model with the addition of a minimum entropy loss outperforms all compared approaches on all classes, thus improving on the known state-of-the-art in this area.

\section{Conclusion}
In this paper, we address the problem of domain adaptation for semantic segmentation of aerial images using entropy minimization and class-wise distribution alignment. Our approach improves on the known state-of-the-art performance, and is shown to be efficient in addressing domain shift on a challenging aerial dataset. We demonstrate the benefit of applying soft class-wise distribution alignment which results in improved performance compared with existing work. The combination of class-wise distribution alignment and entropy minimization further improves the performance of the proposed approach. In future work, we plan to extend this work to include semi-supervised learning scenarios where a small set of labeled data is available. 



\bibliographystyle{IEEEbib}
\bibliography{ref-org}

\begin{thebibliography}{10}

\bibitem{hoffman2016fcns}
Judy Hoffman, Dequan Wang, Fisher Yu, and Trevor Darrell,
\newblock ``Fcns in the wild: Pixel-level adversarial and constraint-based
  adaptation,''
\newblock {\em CoRR, abs/1612.02649}, 2016.

\bibitem{tsai2018learning}
Yi-Hsuan Tsai, Wei-Chih Hung, Samuel Schulter, Kihyuk Sohn, Ming-Hsuan Yang,
  and Manmohan Chandraker,
\newblock ``Learning to adapt structured output space for semantic
  segmentation,''
\newblock in {\em Proceedings of the IEEE Conference on Computer Vision and
  Pattern Recognition}, 2018, pp. 7472--7481.

\bibitem{vu2019advent}
Tuan-Hung Vu, Himalaya Jain, Maxime Bucher, Matthieu Cord, and Patrick
  P{\'e}rez,
\newblock ``Advent: Adversarial entropy minimization for domain adaptation in
  semantic segmentation,''
\newblock in {\em Proceedings of the IEEE Conference on Computer Vision and
  Pattern Recognition}, 2019, pp. 2517--2526.

\bibitem{li2019bidirectional}
Yunsheng Li, Lu~Yuan, and Nuno Vasconcelos,
\newblock ``Bidirectional learning for domain adaptation of semantic
  segmentation,''
\newblock in {\em Proceedings of the IEEE Conference on Computer Vision and
  Pattern Recognition}, 2019, pp. 6936--6945.

\bibitem{hoffman2018cycada}
Judy Hoffman, Eric Tzeng, Taesung Park, Jun-Yan Zhu, Phillip Isola, Kate
  Saenko, Alexei Efros, and Trevor Darrell,
\newblock ``Cycada: Cycle-consistent adversarial domain adaptation,''
\newblock in {\em International conference on machine learning}, 2018, pp.
  1989--1998.

\bibitem{wu2018dcan}
Zuxuan Wu, Xintong Han, Yen-Liang Lin, Mustafa Gokhan~Uzunbas, Tom Goldstein,
  Ser Nam~Lim, and Larry~S Davis,
\newblock ``Dcan: Dual channel-wise alignment networks for unsupervised scene
  adaptation,''
\newblock in {\em Proceedings of the European Conference on Computer Vision
  (ECCV)}, 2018, pp. 518--534.

\bibitem{benjdira2019unsupervised}
Bilel Benjdira, Yakoub Bazi, Anis Koubaa, and Kais Ouni,
\newblock ``Unsupervised domain adaptation using generative adversarial
  networks for semantic segmentation of aerial images,''
\newblock {\em Remote Sensing}, vol. 11, no. 11, pp. 1369, 2019.

\bibitem{benjdira2020data}
Bilel Benjdira, Adel Ammar, Anis Koubaa, and Kais Ouni,
\newblock ``Data-efficient domain adaptation for semantic segmentation of
  aerial imagery using generative adversarial networks,''
\newblock {\em Applied Sciences}, vol. 10, no. 3, pp. 1092, 2020.

\bibitem{liu2017approximation}
Shuang Liu, Olivier Bousquet, and Kamalika Chaudhuri,
\newblock ``Approximation and convergence properties of generative adversarial
  learning,''
\newblock in {\em Advances in Neural Information Processing Systems}, 2017, pp.
  5545--5553.

\bibitem{chen2018road}
Yuhua Chen, Wen Li, and Luc Van~Gool,
\newblock ``Road: Reality oriented adaptation for semantic segmentation of
  urban scenes,''
\newblock in {\em Proceedings of the IEEE Conference on Computer Vision and
  Pattern Recognition}, 2018, pp. 7892--7901.

\bibitem{hong2018conditional}
Weixiang Hong, Zhenzhen Wang, Ming Yang, and Junsong Yuan,
\newblock ``Conditional generative adversarial network for structured domain
  adaptation,''
\newblock in {\em Proceedings of the IEEE Conference on Computer Vision and
  Pattern Recognition}, 2018, pp. 1335--1344.

\bibitem{chen2019domain}
Minghao Chen, Hongyang Xue, and Deng Cai,
\newblock ``Domain adaptation for semantic segmentation with maximum squares
  loss,''
\newblock in {\em Proceedings of the IEEE International Conference on Computer
  Vision}, 2019, pp. 2090--2099.

\bibitem{shannon1948mathematical}
Claude~E Shannon,
\newblock ``A mathematical theory of communication,''
\newblock {\em Bell system technical journal}, vol. 27, no. 3, pp. 379--423,
  1948.

\bibitem{kullback1951information}
Solomon Kullback and Richard~A Leibler,
\newblock ``On information and sufficiency,''
\newblock {\em The annals of mathematical statistics}, vol. 22, no. 1, pp.
  79--86, 1951.

\bibitem{chen2017deeplab}
Liang-Chieh Chen, George Papandreou, Iasonas Kokkinos, Kevin Murphy, and Alan~L
  Yuille,
\newblock ``Deeplab: Semantic image segmentation with deep convolutional nets,
  atrous convolution, and fully connected crfs,''
\newblock {\em IEEE transactions on pattern analysis and machine intelligence},
  vol. 40, no. 4, pp. 834--848, 2017.

\bibitem{he2016deep}
Kaiming He, Xiangyu Zhang, Shaoqing Ren, and Jian Sun,
\newblock ``Deep residual learning for image recognition,''
\newblock in {\em Proceedings of the IEEE conference on computer vision and
  pattern recognition}, 2016, pp. 770--778.

\bibitem{paszke2017automatic}
Adam Paszke, Sam Gross, Soumith Chintala, Gregory Chanan, Edward Yang, Zachary
  DeVito, Zeming Lin, Alban Desmaison, Luca Antiga, and Adam Lerer,
\newblock ``Automatic differentiation in pytorch,''
\newblock {\em NIPS Workshop}, 2017.

\bibitem{bottou2010large}
L{\'e}on Bottou,
\newblock ``Large-scale machine learning with stochastic gradient descent,''
\newblock in {\em Proceedings of COMPSTAT'2010}, pp. 177--186. Springer, 2010.

\bibitem{labeling2016use}
S~Labeling and B~Vaihingen,
\newblock ``Use of the stair vision library within the isprs use of the stair
  vision library within the isprs 2d,''
\newblock 2016.

\end{thebibliography}


\begin{thebibliography}{10}

\bibitem{cordts2016cityscapes}
Marius Cordts, Mohamed Omran, Sebastian Ramos, Timo Rehfeld, Markus Enzweiler,
  Rodrigo Benenson, Uwe Franke, Stefan Roth, and Bernt Schiele,
\newblock ``The cityscapes dataset for semantic urban scene understanding,''
\newblock in {\em Proceedings of the IEEE conference on computer vision and
  pattern recognition}, 2016, pp. 3213--3223.

\bibitem{everingham2015pascal}
Mark Everingham, SM~Ali Eslami, Luc Van~Gool, Christopher~KI Williams, John
  Winn, and Andrew Zisserman,
\newblock ``The pascal visual object classes challenge: A retrospective,''
\newblock {\em International journal of computer vision}, vol. 111, no. 1, pp.
  98--136, 2015.

\bibitem{lateef2019survey}
Fahad Lateef and Yassine Ruichek,
\newblock ``Survey on semantic segmentation using deep learning techniques,''
\newblock {\em Neurocomputing}, vol. 338, pp. 321--348, 2019.

\bibitem{richter2016playing}
Stephan~R Richter, Vibhav Vineet, Stefan Roth, and Vladlen Koltun,
\newblock ``Playing for data: Ground truth from computer games,''
\newblock in {\em European conference on computer vision}. Springer, 2016, pp.
  102--118.

\bibitem{souly2017semi}
Nasim Souly, Concetto Spampinato, and Mubarak Shah,
\newblock ``Semi supervised semantic segmentation using generative adversarial
  network,''
\newblock in {\em Proceedings of the IEEE International Conference on Computer
  Vision}, 2017, pp. 5688--5696.

\bibitem{wei2018revisiting}
Yunchao Wei, Huaxin Xiao, Honghui Shi, Zequn Jie, Jiashi Feng, and Thomas~S
  Huang,
\newblock ``Revisiting dilated convolution: A simple approach for weakly-and
  semi-supervised semantic segmentation,''
\newblock in {\em Proceedings of the IEEE Conference on Computer Vision and
  Pattern Recognition}, 2018, pp. 7268--7277.

\bibitem{yan2017mind}
Hongliang Yan, Yukang Ding, Peihua Li, Qilong Wang, Yong Xu, and Wangmeng Zuo,
\newblock ``Mind the class weight bias: Weighted maximum mean discrepancy for
  unsupervised domain adaptation,''
\newblock in {\em Proceedings of the IEEE Conference on Computer Vision and
  Pattern Recognition}, 2017, pp. 2272--2281.

\bibitem{long2015learning}
Mingsheng Long, Yue Cao, Jianmin Wang, and Michael Jordan,
\newblock ``Learning transferable features with deep adaptation networks,''
\newblock in {\em International conference on machine learning}, 2015, pp.
  97--105.

\bibitem{long2016unsupervised}
Mingsheng Long, Han Zhu, Jianmin Wang, and Michael~I Jordan,
\newblock ``Unsupervised domain adaptation with residual transfer networks,''
\newblock in {\em Advances in neural information processing systems}, 2016, pp.
  136--144.

\bibitem{hoffman2016fcns}
Judy Hoffman, Dequan Wang, Fisher Yu, and Trevor Darrell,
\newblock ``Fcns in the wild: Pixel-level adversarial and constraint-based
  adaptation,''
\newblock {\em CoRR, abs/1612.02649}, 2016.

\bibitem{tzeng2017adversarial}
Eric Tzeng, Judy Hoffman, Kate Saenko, and Trevor Darrell,
\newblock ``Adversarial discriminative domain adaptation,''
\newblock in {\em Proceedings of the IEEE Conference on Computer Vision and
  Pattern Recognition}, 2017, pp. 7167--7176.

\bibitem{tsai2018learning}
Yi-Hsuan Tsai, Wei-Chih Hung, Samuel Schulter, Kihyuk Sohn, Ming-Hsuan Yang,
  and Manmohan Chandraker,
\newblock ``Learning to adapt structured output space for semantic
  segmentation,''
\newblock in {\em Proceedings of the IEEE Conference on Computer Vision and
  Pattern Recognition}, 2018, pp. 7472--7481.

\bibitem{vu2019advent}
Tuan-Hung Vu, Himalaya Jain, Maxime Bucher, Matthieu Cord, and Patrick
  P{\'e}rez,
\newblock ``Advent: Adversarial entropy minimization for domain adaptation in
  semantic segmentation,''
\newblock in {\em Proceedings of the IEEE Conference on Computer Vision and
  Pattern Recognition}, 2019, pp. 2517--2526.

\bibitem{zou2018unsupervised}
Yang Zou, Zhiding Yu, BVK Vijaya~Kumar, and Jinsong Wang,
\newblock ``Unsupervised domain adaptation for semantic segmentation via
  class-balanced self-training,''
\newblock in {\em Proceedings of the European conference on computer vision
  (ECCV)}, 2018, pp. 289--305.

\bibitem{li2019bidirectional}
Yunsheng Li, Lu~Yuan, and Nuno Vasconcelos,
\newblock ``Bidirectional learning for domain adaptation of semantic
  segmentation,''
\newblock in {\em Proceedings of the IEEE Conference on Computer Vision and
  Pattern Recognition}, 2019, pp. 6936--6945.

\bibitem{hoffman2018cycada}
Judy Hoffman, Eric Tzeng, Taesung Park, Jun-Yan Zhu, Phillip Isola, Kate
  Saenko, Alexei Efros, and Trevor Darrell,
\newblock ``Cycada: Cycle-consistent adversarial domain adaptation,''
\newblock in {\em International conference on machine learning}, 2018, pp.
  1989--1998.

\bibitem{wu2018dcan}
Zuxuan Wu, Xintong Han, Yen-Liang Lin, Mustafa Gokhan~Uzunbas, Tom Goldstein,
  Ser Nam~Lim, and Larry~S Davis,
\newblock ``Dcan: Dual channel-wise alignment networks for unsupervised scene
  adaptation,''
\newblock in {\em Proceedings of the European Conference on Computer Vision
  (ECCV)}, 2018, pp. 518--534.

\bibitem{benjdira2019unsupervised}
Bilel Benjdira, Yakoub Bazi, Anis Koubaa, and Kais Ouni,
\newblock ``Unsupervised domain adaptation using generative adversarial
  networks for semantic segmentation of aerial images,''
\newblock {\em Remote Sensing}, vol. 11, no. 11, pp. 1369, 2019.

\bibitem{benjdira2020data}
Bilel Benjdira, Adel Ammar, Anis Koubaa, and Kais Ouni,
\newblock ``Data-efficient domain adaptation for semantic segmentation of
  aerial imagery using generative adversarial networks,''
\newblock {\em Applied Sciences}, vol. 10, no. 3, pp. 1092, 2020.

\bibitem{kullback1951information}
Solomon Kullback and Richard~A Leibler,
\newblock ``On information and sufficiency,''
\newblock {\em The annals of mathematical statistics}, vol. 22, no. 1, pp.
  79--86, 1951.

\bibitem{liu2017approximation}
Shuang Liu, Olivier Bousquet, and Kamalika Chaudhuri,
\newblock ``Approximation and convergence properties of generative adversarial
  learning,''
\newblock in {\em Advances in Neural Information Processing Systems}, 2017, pp.
  5545--5553.

\bibitem{long2017deep}
Mingsheng Long, Han Zhu, Jianmin Wang, and Michael~I Jordan,
\newblock ``Deep transfer learning with joint adaptation networks,''
\newblock in {\em Proceedings of the 34th International Conference on Machine
  Learning-Volume 70}. JMLR. org, 2017, pp. 2208--2217.

\bibitem{sohn2017unsupervised}
Kihyuk Sohn, Sifei Liu, Guangyu Zhong, Xiang Yu, Ming-Hsuan Yang, and Manmohan
  Chandraker,
\newblock ``Unsupervised domain adaptation for face recognition in unlabeled
  videos,''
\newblock in {\em Proceedings of the IEEE International Conference on Computer
  Vision}, 2017, pp. 3210--3218.

\bibitem{xie2018learning}
Shaoan Xie, Zibin Zheng, Liang Chen, and Chuan Chen,
\newblock ``Learning semantic representations for unsupervised domain
  adaptation,''
\newblock in {\em International Conference on Machine Learning}, 2018, pp.
  5423--5432.

\bibitem{ros2016synthia}
German Ros, Laura Sellart, Joanna Materzynska, David Vazquez, and Antonio~M
  Lopez,
\newblock ``The synthia dataset: A large collection of synthetic images for
  semantic segmentation of urban scenes,''
\newblock in {\em Proceedings of the IEEE conference on computer vision and
  pattern recognition}, 2016, pp. 3234--3243.

\bibitem{chen2017no}
Yi-Hsin Chen, Wei-Yu Chen, Yu-Ting Chen, Bo-Cheng Tsai, Yu-Chiang Frank~Wang,
  and Min Sun,
\newblock ``No more discrimination: Cross city adaptation of road scene
  segmenters,''
\newblock in {\em Proceedings of the IEEE International Conference on Computer
  Vision}, 2017, pp. 1992--2001.

\bibitem{chen2018road}
Yuhua Chen, Wen Li, and Luc Van~Gool,
\newblock ``Road: Reality oriented adaptation for semantic segmentation of
  urban scenes,''
\newblock in {\em Proceedings of the IEEE Conference on Computer Vision and
  Pattern Recognition}, 2018, pp. 7892--7901.

\bibitem{hong2018conditional}
Weixiang Hong, Zhenzhen Wang, Ming Yang, and Junsong Yuan,
\newblock ``Conditional generative adversarial network for structured domain
  adaptation,''
\newblock in {\em Proceedings of the IEEE Conference on Computer Vision and
  Pattern Recognition}, 2018, pp. 1335--1344.

\bibitem{chen2019domain}
Minghao Chen, Hongyang Xue, and Deng Cai,
\newblock ``Domain adaptation for semantic segmentation with maximum squares
  loss,''
\newblock in {\em Proceedings of the IEEE International Conference on Computer
  Vision}, 2019, pp. 2090--2099.

\bibitem{zhu2017unpaired}
Jun-Yan Zhu, Taesung Park, Phillip Isola, and Alexei~A Efros,
\newblock ``Unpaired image-to-image translation using cycle-consistent
  adversarial networks,''
\newblock in {\em Proceedings of the IEEE international conference on computer
  vision}, 2017, pp. 2223--2232.

\bibitem{grandvalet2005semi}
Yves Grandvalet and Yoshua Bengio,
\newblock ``Semi-supervised learning by entropy minimization,''
\newblock in {\em Advances in neural information processing systems}, 2005, pp.
  529--536.

\bibitem{springenberg2015unsupervised}
Jost~Tobias Springenberg,
\newblock ``Unsupervised and semi-supervised learning with categorical
  generative adversarial networks,''
\newblock {\em ICLR}, 2015.

\bibitem{shannon1948mathematical}
Claude~E Shannon,
\newblock ``A mathematical theory of communication,''
\newblock {\em Bell system technical journal}, vol. 27, no. 3, pp. 379--423,
  1948.

\bibitem{chen2017deeplab}
Liang-Chieh Chen, George Papandreou, Iasonas Kokkinos, Kevin Murphy, and Alan~L
  Yuille,
\newblock ``Deeplab: Semantic image segmentation with deep convolutional nets,
  atrous convolution, and fully connected crfs,''
\newblock {\em IEEE transactions on pattern analysis and machine intelligence},
  vol. 40, no. 4, pp. 834--848, 2017.

\bibitem{he2016deep}
Kaiming He, Xiangyu Zhang, Shaoqing Ren, and Jian Sun,
\newblock ``Deep residual learning for image recognition,''
\newblock in {\em Proceedings of the IEEE conference on computer vision and
  pattern recognition}, 2016, pp. 770--778.

\bibitem{paszke2017automatic}
Adam Paszke, Sam Gross, Soumith Chintala, Gregory Chanan, Edward Yang, Zachary
  DeVito, Zeming Lin, Alban Desmaison, Luca Antiga, and Adam Lerer,
\newblock ``Automatic differentiation in pytorch,''
\newblock {\em NIPS Workshop}, 2017.

\bibitem{bottou2010large}
L{\'e}on Bottou,
\newblock ``Large-scale machine learning with stochastic gradient descent,''
\newblock in {\em Proceedings of COMPSTAT'2010}, pp. 177--186. Springer, 2010.

\bibitem{labeling2016use}
S~Labeling and B~Vaihingen,
\newblock ``Use of the stair vision library within the isprs use of the stair
  vision library within the isprs 2d,''
\newblock 2016.

\end{thebibliography}

\end{document}